\documentclass{article}

\usepackage{arxiv}
\usepackage[utf8]{inputenc} 
\usepackage[T1]{fontenc}    
\usepackage{hyperref}       
\usepackage{url}            
\usepackage{booktabs,siunitx}       
\usepackage{amsfonts, amsthm, amsmath, mathtools}       
\usepackage{nicefrac}       
\usepackage{microtype}      
\usepackage{graphicx}
\usepackage{natbib}
\usepackage{algorithm, algpseudocode, doi}
\usepackage{thmtools, thm-restate, cleveref}
\usepackage[toc,page]{appendix}
\usepackage{pgfplots, mathrsfs}
\pgfplotsset{compat=1.15}
\usetikzlibrary{arrows}

\newcommand{\ep}{\epsilon}

\newcommand{\var}{\text{Var}}

\theoremstyle{definition}
\newtheorem{example}{Example} 
\newtheorem{theorem}{Theorem}
\newtheorem{lemma}[theorem]{Lemma} 
\newtheorem{proposition}[theorem]{Proposition} 
\newtheorem{remark}[theorem]{Remark}

\newtheorem{definition}[theorem]{Definition}

\title{Optimal Survey Design for Private Mean Estimation}

\author{
    \href{https://orcid.org/0000-0000-0000-0000}{\includegraphics[scale=0.06]{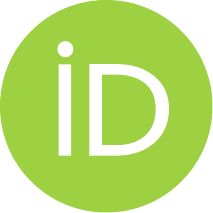}\hspace{1mm}
    Yu-Wei Chen}\\
	Department of Statistics\\
	Purdue University\\
	West Lafayette, IN 47907 \\
	\texttt{chen4357@purdue.edu} \\
    \And
	\href{https://orcid.org/0000-0000-0000-0000}{\includegraphics[scale=0.06]{orcid.pdf}\hspace{1mm}
    Raghu Pasupathy}\\
	Department of Statistics\\
	Purdue University\\
	West Lafayette, IN 47907 \\
	\texttt{pasupath@purdue.edu} \\
    \And
	\href{https://orcid.org/0000-0000-0000-0000}{\includegraphics[scale=0.06]{orcid.pdf}\hspace{1mm}
    Jordan A. Awan}\\
	Department of Statistics\\
	Purdue University\\
	West Lafayette, IN 47907 \\
	\texttt{jawan@purdue.edu} \\
}

\begin{document}
\maketitle

\begin{abstract}
    This work identifies the first privacy-aware stratified sampling scheme that minimizes the variance for general private mean estimation under the Laplace, Discrete Laplace (DLap) and Truncated-Uniform-Laplace (TuLap) mechanisms within the framework of differential privacy (DP). 
    We view stratified sampling as a subsampling operation, which amplifies the privacy guarantee; however, to have the same final privacy guarantee for each group, different nominal privacy budgets need to be used depending on the subsampling rate. 
    Ignoring the effect of DP, traditional stratified sampling strategies risk significant variance inflation.
    We phrase our optimal survey design as an optimization problem, where we determine the optimal subsampling sizes for each group with the goal of minimizing the variance of the resulting estimator. 
    We establish strong convexity of the variance objective, propose an efficient algorithm to identify the integer-optimal design, and offer insights on the structure of the optimal design.
\end{abstract}

\section{Introduction} \label{sec: Intro}

Differential Privacy (DP), introduced by \citet{dwork2006calibrating}, is a popular probabilistic framework designed to protect individual privacy while preserving the utility of data. By introducing calibrated random noise into the data processing, DP ensures that outputs remain informative while reducing the risk of identifying individuals. However, this added noise introduces unique challenges for data analysis. Neglecting the effects of DP mechanisms can lead to biased and incorrect conclusions (\citealp{Alexis2020}; \citealp{Christopher2021}). To address these challenges, researchers commonly employ various inference strategies, such as Bayesian inference (\citealp{Bernstein2018} $\&$ \citeyear{bernstein2019differentially}; \citealp{schein2019locally}; \citealp{kulkarni2021differentially}; \citealp{ju2022data}), asymptotic analysis (\citealp{gaboardi2016differentially}; \citealp{gaboardi2017local}; \citealp{wang2018statistical}), simulation-based inference \citep{awan2024simulation}, and bootstrapping methods (\citealp{ferrando2022parametric}; \citealp{wang2022differentially}). However, there is also a growing need to integrate DP into the design of data collection schemes.

Survey sampling traditionally encompasses three components: sample selection, data collection, and estimation \citep{brick2011}. Over time, survey sampling has been evolving to incorporate new technologies \citep{frankel1987fifty}, such as registration-based sampling \citep{green2006can}, telephone sampling \citep{force2010new}, and computerization \citep{baker1998computer}. Differential privacy represents one of the latest advances, fostering the need to optimize data collection to balance privacy and utility.

Among survey sampling methods, stratified sampling stands out as a robust scheme that leverages auxiliary information to collect valuable samples that can minimize variance. Unlike simple random sampling (SRS), stratified sampling minimizes the risk of having bad samples by dividing the population into groups (strata) based on common characteristics \citep{lohr2021sampling}. \citet{neyman1934two} was the first to formalize stratified sampling, introducing an optimal allocation of samples to minimize variance across groups---a goal aligned with the principles of experimental design \citep{wu2011experiments}.

Developing DP techniques for surveys is very important to protect individual respondents, especially when sensitive questions are asked. Furthermore, another key motivation for incorporating DP is as a technique to reduce response bias---also known as answer bias---which often arises when individuals avoid answering sensitive or controversial questions truthfully, leading to skewed or inaccurate conclusions. Randomized response, introduced by \citet{Stanley1965}, provides a mechanism for respondents to address such questions while satisfying differential privacy \citep{dwork2014algorithmic}. Incorporating appropriate noise through DP techniques, our framework effectively balances data utility with individual privacy and can also reduce response biases.

When considering differential privacy for survey sampling, it is first important to recognize a crucial result of differential privacy under sampling which plays a key role in the formulation of our problem: When a privacy mechanism is applied to a randomly sampled subset of a population (while the sampled individuals themselves remain secret), a stronger privacy guarantee can be achieved \citep{kasiviswanathan2011can}. This effect is referred to as the ``secrecy of the sample'' or privacy amplification by subsampling. Thus, in stratified sampling, where the population is divided into subpopulations, subsampling within groups can amplify privacy protection. This effect adds complexity to the optimization problem of determining the optimal survey design when integrating differential privacy into stratified sampling.

This paper is the first to consider optimizing a survey design when incorporating a differential privacy guarantee. Specifically, we develop an optimal stratified sampling scheme to minimize the estimator variance in private mean estimation under differential privacy. Holding the total sample size fixed, we search for the optimal subgroup sizes. A key challenge is that different subsampling rates for each group require different ``nominal privacy budgets'' in order to give the same privacy guarantee to all members of the population, which results in a complex objective function.  Ignoring DP-induced variance during the design phase can cause significant inflation in estimator variance, as demonstrated in Section~\ref{sec: Simulation} and highlighted in the table below. Table~\ref{tab: variance ratio demo} illustrates this issue by comparing the variance ratio of the naive design to the optimal design, using both Laplace and Truncated-Uniform-Laplace (TuLap) mechanisms \citep{Awan_Slavković_2018}.

\begin{table}[H]
    \begin{center}
       \begin{tabular}{llllll}
          \toprule
          {$\epsilon$} & {0.1} & {$10^{-1/2}$} & {1} & {$10^{1/2}$} & {10} \\ 
          \midrule
          Laplace & 1.828 & 2.095 & 2.269 & 2.311 & 1.973 \\ 
          TuLap & 2.405 & 3.324 & 3.877 & 4.06 & 4.076 \\ 
          \bottomrule
       \end{tabular}
    \end{center}
   \caption{Variance ratio on population mean}
   \label{tab: variance ratio demo}
\end{table}

\textbf{Contributions:} We propose a novel framework for designing stratified sampling schemes under a hybrid local/central differential privacy regime, leveraging a design of experiment (DOE) perspective. We provide an efficient algorithm for locating the optimal integer design, which enables practitioners to conveniently implement in practice. Our approach accounts for variance from sampling as well as from the DP noise amplified by subsampling during the design phase, allocating the best subsampling sizes to minimize the variance of our final estimator.

This work is, to the best of our knowledge, the first to apply experimental design principles to data collection under differential privacy, fundamentally altering optimal stratified sampling schemes to accommodate DP considerations. Our contributions include the following:
\begin{itemize}
    \item  We identify and formulate the problem as a constrained integer-programming problem, identifying its alignment within the framework of DOE.
    \item  We establish strong convexity for a general variance objective, covering important cases such as A-optimality (minimizing the trace of the covariance matrix) and population mean estimation, under three common additive DP mechanisms: Laplace, Discrete Laplace, and Truncated-Uniform-Laplace.
    \item  For the population mean estimate, we derive closed-form continuous solutions under Discrete Laplace and Truncated-Uniform-Laplace mechanisms; additionally, we derive the optimal continuous design when using purely Laplace noise, which reveals the DP-aware design lies between the original, no-noise design and the pure Laplace noise design.
    \item  By leveraging the strong convexity of the variance objectives, we develop a computationally efficient algorithm to locate the integer-optimal design, overcoming the intractability of exhaustive search methods.
\end{itemize}

\textbf{Organization:} The rest of the paper is structured as follows: Section~\ref{sec: Background} provides the necessary background on local and central differential privacy as well as privacy amplification by subsampling. Section~\ref{sec: problem setup} formulates the main problem, a convex-constrained minimization problem with a general variance objective. Section~\ref{sec: theory and alg.} establishes the strong convexity of the problem, a key property enabling the efficient search for the integer-optimal design. In Section~\ref{sec: Simulation}, we illustrate variance inflation resulting from naive stratified sampling without considering DP effects and demonstrate the efficiency of our algorithm in locating the integer-optimal design, even as the number of groups increases. Finally, Section~\ref{sec: discussion} discusses the implications of our findings and potential avenues for future work.

\textbf{Related Work:} Although optimal survey design for private estimation remains unexplored, differentially private survey sampling has recently been studied in other contexts. \citet{lin2024differentially} construct confidence intervals for proportions using data collected through stratified sampling. \citet{bun2020controlling} examine stratified and cluster sampling, highlighting that certain sampling schemes can degrade privacy rather than enhance it and can increase privacy risks.

Sampling has also been employed as a technique to address DP-related problems. \citet{ebadi2016sampling} examine the impact of various sampling schemes on differential privacy, demonstrating that only Bernoulli sampling amplifies privacy protection. \citet{joy2017differential} propose a sampling-based privacy mechanism satisfying differential privacy, while \citet{bichsel2018finder} develop a correlated sampling method to detect privacy violation. 

The concept of ``secrecy of the sample,'' proposed and formalized by \citet{kasiviswanathan2011can}, highlights the role of random sampling from population in enhancing privacy guarantees while keeping the members of the dataset secret. \citet{li2012sampling} demonstrate that implementing $k-$anonymity safely after a random sampling step ensures $(\epsilon,\delta)-$DP. \citet{cheu2019distributed} employ ``secrecy of the sample'' to establish the privacy guarantee of the shuffled model, an intermediate variant between central and local models that enhances privacy by relying on a trusted curator to shuffle the locally privatized data before releasing a final private statistic. \citet{heber2021random} incorporate random sampling into their solution for multivariate frequency estimation in locally differentially private (LDP) settings.

Variance minimization and estimation, as well as utility-maximization mechanisms, have been widely investigated under DP. Treating multi-agent systems as probabilistic models of environmental states parameterized by agent profiles, \citet{wang2017dpminvariance} establish a lower bound on the $l_1$-induced norm of the covariance matrix for minimum-variance unbiased estimators when the agents’ profiles are $\epsilon$-DP. \citet{li2023fine} expose how output poisoning attacks can manipulate and deteriorate mean and variance estimation under local DP. \citet{amin2019bounding} identify a bias-variance trade-off caused by clamping in DP learning and provide careful tuning on the clamping bound. For a fixed count query, \citet{ghosh2009universally} show that the geometric mechanism minimizes the expected loss for virtually all possible users while satisfying the DP constraint. Similarly, for a single real-valued query function, \citet{geng2015optimal} demonstrate that the staircase mechanism can minimize $\ell_1$ and $\ell_2$ costs under specific parameter settings.

\section{Background} \label{sec: Background}

We introduce the necessary background of local and central differential privacy and relevant subsampling results.

Differential privacy can be ensured from two perspectives: local DP and central DP. Both approaches achieve privacy guarantees by employing randomized mechanisms that perturb sensitive data or statistics and produce their privatized outputs. Local DP offers stronger privacy protection by privatizing individual data, ensuring that sensitive information remains unknown to anyone, thereby shielding individuals from both internal and external threats. However, this comes at the cost of reduced data utility. In contrast, central DP relies on trusted data curators to collect sensitive data and subsequently release a privatized summary, protecting individual only from external adversaries. 

\begin{definition}[Local Differential Privacy: \citet{duchi2013local}]
    Let $\mathcal{X}$ be the set of possible contributions of an individual. A randomized privacy mechanism ${M}$ provides local $\epsilon-$differential privacy, if for any two data points $x, x' \in \mathcal{X}$ and for any measurable set $S \subseteq \text{Range}({M})$.
    \begin{align*}
        \Pr[{M}(x) \in S] \leq e^{\epsilon} \Pr[{M}(x') \in S].
    \end{align*}
     
\end{definition}

To safeguard individual privacy through added noise, the amount of noise must be carefully quantified, with sensitivity playing a pivotal role in this process. Greater data dispersion increases sensitivity, which in turn requires scaling up the noise in the privacy mechanism.

\begin{definition}[Sensitivity]
    Let $f: \mathcal{X} \to \mathbb{R}$ be a statistic. The sensitivity of $f$ is 
        $\Delta f = \max_{\substack{%
    x, x' \in \mathcal{X}\\
     }}
      \!  | f(x) - f(x') |$.
\end{definition}

In Example~\ref{ex: epsilon DP mechanisms}, we introduce three differentially private mechanisms to which we apply our findings throughout the paper.

\begin{example} \label{ex: epsilon DP mechanisms}
The following are three common DP mechanisms. 
For local $\epsilon-$DP, given a real-valued statistic $f(x)$,
    \begin{itemize}
    \item the Laplace mechanism is $Z = f(x) + L$, where $L \sim \text{Lap}(0,s)$ with $s =\Delta f/\epsilon$.
    \item the Discrete Laplace mechanism (DLap) \citep{Inusah_Kozubowski_2006, ghosh2009universally} for integer-valued data is $Z = f(x) + K$, where $K \sim P(K = k) = \frac{1-p}{1+p} \ p^{|k|}$ with $p = \exp{\left( \frac{\epsilon}{\Delta f} \right)}$.
    \item the Truncated-Uniform-Laplace mechanism (TuLap) \citep{Awan_Slavković_2018} is $Z = f(x) + K + U$, where $K$ is the same as in DLap and $U \sim \text{Uniform}(-\frac{1}{2}, \frac{1}{2})$. TuLap is a canonical noise distribution \citep{awan2023canonical} and it is related to the Staircase distribution \citep{geng2015optimal}.
    \end{itemize}    
\end{example}

While local DP offers strong protection against both external adversaries as well as the data collectors themselves, it requires a large amount of noise for privatization. For example, \citet{duchi2013local} show that local DP mechanisms have inferior asymptotic variance compared to non-private estimators. On the other hand, central DP has a trusted curator, but gives the same DP guarantee to external adversaries and allows for asymptotically negligible noise to be added \citep{smith2011privacy-preserving, barber2014privacy}.

\begin{definition}[Central Differential Privacy: \citet{dwork2006calibrating}]
    Let $\mathcal{X}^n$ be the set of possible datasets with sample size $n$ and $d_H(\cdot, \cdot)$ be the Hamming distance on $\mathcal{X}^n \times \mathcal{X}^n$, a randomized privacy mechanism ${M}$ provides (central) $\epsilon-$differential privacy, if for any two  datasets $X, X' \in \mathcal{X}^n$ such that $d_H(X, X') \leq 1$, and for any measurable set $S \subseteq \text{Range}({M})$,
    \begin{align*}
        \Pr[{M}(X) \in S] \leq e^{\epsilon} \Pr[{M}(X') \in S].
    \end{align*}
\end{definition}

Note that all local DP mechanisms are also central DP. Therefore, the following lemmas on central DP can be applied to local DP mechanisms.  

\begin{lemma}[Parallel Composition: \citet{mcsherry2009privacy}] \label{lemma: parallel composition}
    Let $M_1, M_2, \dots, M_k$ be a set of $k$ mechanisms, where each $M_i$ satisfies $\epsilon_i-$DP. Suppose these mechanisms are applied to disjoint subsets of the dataset $\mathcal{X}^n$, denoted as $D_1, D_2, \dots, D_k$, such that $\mathcal{X}^n = \bigcup_{i=1}^k D_i$. Assume that the sizes of $D_i$'s are public. Then, the combined mechanism $M = (M_i)_{i=1}^k$ satisfies $\max_i \epsilon_i-$DP.
\end{lemma}

Lemma~\ref{lemma: parallel composition} states that the ultimate privacy guarantee of a set of privacy mechanisms applying to disjoint datasets only hinges on the worst among all guarantees. In stratified sampling, a set $D_i$ represents a stratum (group).

\begin{lemma}[Subsampling: Corollary 3, \citealp{dong2022gaussian}; \citealp{ullman2017}] \label{lemma: subsampling operator}
If $M$ is a privacy mechanism that satisfies $\ep$-DP for a dataset of size $n$, and $S_m$ is the subsampling operator that chooses a subset of size $m$ from the dataset of size $n$ uniformly at random, then $M \circ S_m$ satisfies $\log(1-q+q\exp(\ep))-$DP, where $q=m/n$.
\end{lemma}

Lemma~\ref{lemma: subsampling operator} shows how subsampling creates its randomness, thereby bringing about privacy amplification. It immediately follows from Lemma~\ref{lemma: parallel composition} that if $N$ is decomposed into $k$ disjoint groups of size $N_i$ for $i=1,\ldots, k$, and we want to sample subsets of size $n_i$ from each group, uniformly at random, then the privacy guarantee for a nominal $\ep$-DP mechanism $M$ applied to the subsamples is $\log(1-q_{\max}+q_{\max}\exp(\ep))$, where $q_{\max}=\max_i \frac{n_i}{N_i}$. 

Finally, we recall the post-processing property of DP: If ${M}: \mathcal{X}^n \to \mathcal{Y}$ is an $\epsilon-$DP mechanism and $g: \mathcal{Y} \to \mathcal{Z}$ is another mechanism, then $g \circ {M}: \mathcal{X}^n \to \mathcal{Z}$ satisfies $\epsilon-$DP \citep{dwork2014algorithmic}. This property allows us to construct customized estimators from the DP outputs, without compromising the privacy guarantee.

\section{Problem Setup} \label{sec: problem setup}

In this paper, we minimize the variance of a mean estimator, which comprises data randomness and additive privacy noise centered at zero. Our method has dual privacy guarantees: a local DP guarantee from the nominal privacy loss budget against the data collector, and a central DP guarantee, boosted by subsampling, against external adversaries. 

Suppose there are $k$ groups (strata) of people $D_i$ with size $N_i$, $i=1,\ldots,k$, that make up the entire population. In each group $i$, $Y_{ij}$ represents the feature of the $j-$th individual, $j=1,\ldots,N_i$, which has mean $\mu_i$ and variance $\sigma_i^2$ with bounded support. In a local DP setting, instead of $Y_{ij}$, $Z_{ij} = Y_{ij} + W_{ij}$ is the privatized response one can access, where $W_{ij}$ is the i.i.d. additive noise with mean $0$ and finite variance $\gamma^2$ depending on $n_i, N_i$ and $\epsilon$. Since we can only draw $n_i$ samples from group $D_i$ with a total sample size of $\eta$, the constrained minimization problem of interest becomes
\begin{align} \label{min: general}
    \arg\min& \sum_{i=1}^k \frac{\alpha_i^2}{n_i} \left[ \sigma_i^2 + \gamma^2 (n_i, N_i, \epsilon) \right],
\end{align}
subject to  $\sum n_i = \eta$, where $n_i$ is searched over $\mathbb{N}$,  $\alpha_i$ is weight, and $\eta$ is a pre-determined total sample size.

This problem is classified as nonlinear integer programming, where both the objective and the constraint are convex; specifically, the variance objective is strongly convex (Theorem~\ref{thm: strong convexity}). As it will be addressed later using the Lagrangian, which introduces a continuous multiplier for the equality constraint, it can be generally treated as a mixed-integer programming problem \citep{lee2011mixed}.

We assume $\alpha_i$, $N_i$, $\sigma_i$, $(i=1,..,k)$ and $\eta$ are given or determined prior to subsampling. The following examples show how the $\alpha_i$ can be chosen to optimize for various variance objectives. 
\begin{example}[Population Mean Estimation]
    One of the most important parameters to estimate in survey sampling is the population mean. For group $i$, the group mean $\mu_i$ can be estimated by 
        $\hat{\mu}_i = \frac{1}{n_i} \sum_{j=1}^{n_i} Z_{ij}$, 
    which is an unbiased estimator of $\mu_i$. The population mean $\mu$ can thus be unbiasedly estimated by 
        $\hat{\mu} = \frac{\sum_{i=1}^k N_i \hat{\mu}_i}{\sum_{i=1}^k N_i}$. 
    Then, $\var (\hat{\mu}) = \frac{1}{(\sum N_i)^2} \sum_{i=1}^k \frac{{N_i}^2}{n_i} \left[ \sigma_i^2 + \gamma^2 (n_i, N_i, \epsilon) \right]$. Thus, the constrained minimization problem is:
    \begin{align} \label{min: population variance objective}
        {\arg\min}& \sum_{i=1}^k \frac{{N_i}^2}{n_i} \left[ \sigma_i^2 + \gamma^2 (n_i, N_i, \epsilon) \right] \\
        \text{s.t.}& \sum n_i = \eta, \quad n \in \mathbb{N}^k. \nonumber
    \end{align}
    This aligns with \eqref{min: general} as $\alpha_i = N_i$ for all $i$.
\end{example}

\begin{example}[A-Optimal Experimental Design]
     Another important case is the A-optimal experimental design, where we minimize the trace of the covariance matrix subject to the constraint. Denote the covariance matrix of $(\hat{\mu}_1,\ldots,\hat{\mu}_k)^{\top}$ as $E_{\mu}$, then the A-optimal experimental design is
    \begin{align} \label{min: A-optimal objective}
        \arg \min& \ \text{tr}(E_{\mu}) = \sum_{i=1}^k \frac{1}{n_i} \left[ \sigma_i^2 + \gamma^2 (n_i, N_i, \epsilon) \right] \\
        \text{s.t.}& \sum n_i = \eta, \quad n \in \mathbb{N}^k. \nonumber
    \end{align}
    This algins with \eqref{min: general} as $\alpha_i = 1$ for all $i$.
\end{example}

\begin{remark}
    There are other interesting settings that fit into this framework. For instance, one may be interested in a unit-free optimal design by setting $\alpha_i=1/\sigma_i$.
\end{remark}

Now, we look into the variance component from the privacy mechanism, that is, $\gamma^2 (n_i, N_i, \epsilon)$ for all $i$, where the subsampling comes into play. Rather than using $q_{\max}$ for all $k$ groups as in Lemma~\ref{lemma: subsampling operator}, we consider $q_i = \frac{n_i}{N_i}$ such that $M_i\circ S_{n_i}$ satisfies $\epsilon-$DP, where $M_i$ is the privacy mechanism for group $i$ and $S_{n_i}$ is the subsampling operator which chooses a subset of size $n_i$ from the group $i$. This ensures every person gets the same level of central DP privacy protection, regardless of their group size. 

\begin{proposition} \label{prop: nominal epsilon for epsilon-DP}
    If the nominal privacy budget of $M_i$ is $\log \left( \frac{\exp{(\epsilon/\Delta f)} - 1 + q_i}{q_i} \right)$, then $M_i\circ S_{n_i}$ satisfies $\epsilon-$DP. 
\end{proposition}

Proposition~\ref{prop: nominal epsilon for epsilon-DP} establishes the dual privacy guarantees for our mechanisms, ensuring uniform privacy protection for all individuals from public disclosure.

\begin{example} \label{eg: general variance from the three mechanisms}
    Applying Proposition~\ref{prop: nominal epsilon for epsilon-DP} to the three mechanisms, we have the following, where $s_i = 1 / {\log \left( 1 + \frac{\exp{(\epsilon / \Delta f)} - 1}{n_{i}} N_i \right)}$:
    \begin{itemize}
        \item The Laplace mechanism for local $\epsilon-$DP is $Z_{ij} = Y_{ij} + s_i \cdot \text{Lap}(0,1)$.
            Then, the variance objective becomes
            \begin{align} \label{stat: general variance from Laplace}
                \hspace{-.6cm}\sum_{i=1}^k \frac{{\alpha_i}^2}{n_i} \left[ \sigma_i^2 + 2 \log^{-2} \left( 1+ \frac{\exp{(\epsilon / \Delta f)}-1}{n_i} N_i \right) \right].
            \end{align}
        \item The Discrete Laplace mechanism for local $\epsilon-$DP is $Z_{ij} = Y_{ij} + K_{ij}$, where 
            $K_{ij} \sim \text{DLap} \left( p_i = \exp(1/s_i)\right)$. 
        Then, the variance objective becomes
        \begin{align} \label{stat: general variance from DLap}
            \sum_{i=1}^k \frac{{\alpha_i}^2}{n_i} \left[ \sigma_i^2 + 2 \frac{ \frac{n_i}{N_i} (\exp{(\epsilon / \Delta f)}-1+\frac{n_i}{N_i})}{(\exp{(\epsilon / \Delta f)}-1)^2} \right].
        \end{align}
        \item The Truncated-Uniform-Laplace mechanism for local $\epsilon-$DP is $Z_{ij} = Y_{ij} + K_{ij} + U_{ij}$
        where $K_{ij} \sim \text{DLap} \left( p_i = \exp(1/s_i)\right)$. 
        and $U_{ij} \sim \text{Uniform}(-\frac{1}{2}, \frac{1}{2})$.
        Then, the variance objective becomes
        \begin{align} \label{stat: general variance from TuLap}
            \sum_{i=1}^k \frac{{\alpha_i}^2}{n_i} \left[  \sigma_i^2 + \frac{1}{12} + 2 \frac{\frac{n_i}{N_i} (\exp{(\epsilon / \Delta f)}-1+\frac{n_i}{N_i})}{(\exp{(\epsilon / \Delta f)}-1)^2} \right].
        \end{align}
    \end{itemize}    
\end{example}

By using an exhaustive search, the complexity of solving \eqref{min: general} is $\binom{\eta-1}{k-1}$ \citep{Ross1974}, motivating the need for a customized optimization method.

\section{Theoretical Results and Algorithm} \label{sec: theory and alg.}

In this section, we develop the optimal integer design which solves \eqref{min: general}, introduced in Secion~\ref{sec: problem setup}. In Section~\ref{sec: strong convexity}, we discover and prove the strong convexity of the variance objectives under Laplace, DLap, and TuLap mechanisms, which enables us to precisely locate the optimal design in Section~\ref{subsec: algorithm}. In Section~\ref{subsec: closed form solution} we also derive some closed-form solutions over continuous space for some special cases.

\subsection{Strong Convexity} \label{sec: strong convexity}

We begin by establishing the strong convexity of our variance objectives. Strong convexity ensures a unique optimum over the reals and provides a quadratic lower bound for the objective function. These properties are leveraged in Section~\ref{subsec: algorithm} to develop an efficient optimization algorithm.

\begin{restatable}[Strong Convexity]{thm}{strongConvexity} \label{thm: strong convexity}
Let $\alpha_i$, $N_i$, $\sigma_i$, and $\eta$ be given for all $i=1,\ldots, k$. 
        Then, the continuous relaxations of the variance objectives \eqref{stat: general variance from Laplace}, \eqref{stat: general variance from DLap} and \eqref{stat: general variance from TuLap}, with $(n_1,\ldots,n_k)$ replaced with $(x_1,\ldots,x_k) \in \mathbb{R}^k_+$ such that $\sum x_i = \eta$, are strongly convex.
\end{restatable}

\begin{proof}[Proof Sketch]
    We first prove that the variance objectives are strongly convex over $(0,\eta)^k$ and then show that this property continues to hold under the convex constraint. 
    For the DLap and TuLap mechanisms, strong convexity follows by inspection; for Laplace, strong convexity is established by change of variables and successive differentiation.    
\end{proof}

Strong convexity in Theorem~\ref{thm: strong convexity} guarantees a unique solution over the reals. A straightforward approach to solving it is using Newton's method with established \texttt{R} packages such as \texttt{optim}, \texttt{nloptr} or \texttt{alabama}. However, solving the mixed-integer programming problem is more complex, as existing packages do not provide direct solutions. Strong convexity is key in identifying the integer-optimal design in Section~\ref{subsec: algorithm}. Note that neither CVX’s free solvers in \texttt{MatLab} nor any \texttt{R} package support mixed-integer programming. 


    

\subsection{Closed-Form Solution for Population Mean} \label{subsec: closed form solution}

A closed-form continuous solution is desirable for its ease of implementation and the insights it provides into the behavior of the design. While such a solution does not exist for all $\alpha_i$, intriguing results emerge when $\alpha_i = N_i$, the case of population mean estimation.

\begin{restatable}[Closed-Form Solutions for Population Mean]{prop}{closedFormSolution} \label{prop: solution of DP variance from DLap and TuLap}
    If $M_i$ is Discrete Laplace or TuLap and $\alpha_i = N_i$ for all $i$, the continuous solution of \eqref{stat: general variance from DLap} and \eqref{stat: general variance from TuLap} under the constraint $\{ (x_1,\ldots,x_k) \in \mathbb{R}_{+}^k: \eta = \sum x_i \}$ have a closed form  
        $x_i^* = [(\tau_i N_i)/(\sum \tau_i N_i) ]\eta$,  
    where $\tau_i^2 = \sigma_i^2$ for Discrete Laplace and $\tau_i^2 = \sigma_i^2 + \frac{1}{12}$ for TuLap ($x_i^*$ plays the role of $n_i^*$).
\end{restatable}

\begin{proof}[Proof Sketch] 
    We first formulate the Lagrangian of the constrained optimization problem. The KKT conditions give a proportional relation $x_i \propto \tau_i N_i$ for all $i$. Then, the constraint provides a unique solution for the $x_i$'s.
\end{proof}

\begin{remark}
The solution of \eqref{stat: general variance from DLap} is identical to the naive stratified sampling design, also known as the Neyman allocation \citep{neyman1934two} or the optimal allocation \citep{kempf2004encyclopedia}, where the sample size allocated to each group is proportional to both the variability and the size of the group. In contrast, \eqref{stat: general variance from TuLap} has a regularization effect on its sample sizes that results in a different allocation.
\end{remark}
    
As for the Laplace noise, although we were unable to derive a general closed-form solution for \eqref{stat: general variance from Laplace}, we have an interesting finding in the case of population mean estimates, which offers insight in the interplay between the no-DP and purely DP solutions. First, we define the non-private variance as 
\begin{align} \label{stat: original variance}
    \frac{1}{(\sum N_i)^2} \left\{ \sum_{i=1}^k \frac{{N_i}^2}{x_i} \sigma_i^2 \right\}.
\end{align}
Under the constraint of  $C = \{ \mathrm{x} \in \mathbb{R}^k : \sum x_i = \eta\}$, the non-private variance \eqref{stat: original variance} is minimized at 
        $x_i^* = (\sigma_i N_i/\sum \sigma_i N_i) \eta,$ 
for all $i$. On the other hand, we define the pure DP variance as 
\begin{align} \label{stat: pure DP variance from Laplace}
    \frac{1}{(\sum N_i)^2} \left\{ \sum_{i=1}^k \frac{{N_i}^2}{x_i} \left[ 2 \log^{-2} \left( 1+ \frac{\exp{(\epsilon / \Delta f)}-1}{x_i/N_i} \right) \right] \right\}.
\end{align}

\begin{restatable}[Closed-Form Solution of Purely Laplace Variance]{prop}{closedFormSolutionPureDP} \label{prop: solution of pure DP variance from Laplace}
Under the constraint of  $C = \{ x \in \mathbb{R}^k : \sum x_i = \eta\}$, the pure DP variance from the Laplace mechanism \eqref{stat: pure DP variance from Laplace} is minimized at 
        $x_i^* = (N_i/\sum N_i) \eta$, 
for all $i$.
\end{restatable}

\begin{proof}[Proof Sketch]
    In addition to the similar Lagrangian proof argument in Proposition~\ref{prop: solution of DP variance from DLap and TuLap}, we use the result that $\varphi (y) = \frac{2}{y} \log^{-2} \left( 1+ \frac{c}{y} \right)$ is strongly convex, as proved in Theorem~\ref{thm: strong convexity}, to identify the solution form.
\end{proof}

    Proposition~\ref{prop: solution of pure DP variance from Laplace} indicates that the sample size allocated to each group is merely proportional to the size of the group. This corresponds to the concept of the proportional allocation \citep{kempf2004encyclopedia}.  It is surprising that the solution is independent of $\epsilon$ as \eqref{stat: pure DP variance from Laplace} can be blown up when $\epsilon$ drops. Although $\epsilon$ does not play a role in either the solution of the original variance or that of the pure DP variance, it plays a critical role in the solution of the total variance. We illustrate this phenomenon in Section~\ref{sec: Simulation}.

\subsection{Algorithm to Find the Optimal Integer Design} \label{subsec: algorithm}

Since Theorem~\ref{thm: strong convexity} only ensures the existence and uniqueness of the continuous-optimal design, in this section we use the strong convexity property to derive a small region that is guaranteed to contain the integer-optimal design. This result is leveraged in Algorithm~\ref{alg: integer-optimal design} to efficiently find the integer-optimal design.

The search over integer points satisfying the convex constraint is finite, as the set of feasible integer-valued solutions, $D = \{ n \in \mathbb{N}^k : \sum n_i = \eta\}$, is inherently limited. Moreover, the strong convexity of the variance objective provides a quadratic lower bound, ensuring that a certain level set of this bound must contain the integer-optimal point. This level set can be characterized in terms of Euclidean distance, with the radius determined by the smallest eigenvalue of the objective function.

\begin{lemma}[Range to Search for Integer-Optimal Design] \label{lemma: range to search}
    Let $g_1, g_2, g_3$ be the objective function  from \eqref{stat: general variance from Laplace}, \eqref{stat: general variance from DLap}$, \eqref{stat: general variance from TuLap}$ respectively, and $C = \{ x \in \mathbb{R}^k: \sum_{i=1}^k x_i = \eta, \ x>0\}$ such that $x^* = \arg \min_C g_j (x)$, then
    \begin{align} \label{eq: integer-optimal design}
        n^* = \arg \min_D g_j (n),
    \end{align}
    is located within $\overline{{B_{x^*}(r)}} = \{ x: \|x-x^*\|_2 \leq r \}$ with $r=\sqrt{2 (g_j (n_{\text{init.}}) - g_j (x^*))/\lambda}$ for any given $j \in \{1,2,3\}$, where $\lambda$ is the smallest eigenvalue of the Hessian of $g_j(x^*)$ and $n_{\text{init.}} = \arg \min_{E} g_j(n)$ with   $E = \{ n \in \mathbb{N}^k : \sum n_i = \eta, \ \left\lfloor x_i^*  \right\rfloor \leq n_i \leq \left\lceil x_i^* \right\rceil \}\}$.
\end{lemma}

Applying the result of Lemma~\ref{lemma: range to search}, we propose Algorithm \ref{alg: integer-optimal design} which starts with the continuous solution, identifies a small set of candidate integer solutions, and then identifies the integer-optimal design within this smaller set.

\algrenewcommand\algorithmicrequire{\textbf{Input:}}
\algrenewcommand\algorithmicensure{\textbf{Output:}}

\begin{algorithm}
\caption{Integer-Optimal Design} \label{alg: integer-optimal design}
\begin{algorithmic}
\Require $x^*$ (the optimal continuous solution) and Hessian matrix of $g: H_g (x^*)$
    \For{$i = 1,...,k-1$}
        \State Define $T_i = \{ n_i \in \mathbb{N}: \left\lfloor x_i^*  \right\rfloor \leq n_i \leq \left\lceil x_i^* \right\rceil \}$
    \EndFor
    \State Define $T = \{ (n_1,...,n_{k-1}, n_k): n_k = \eta - \sum_{i=1}^{k-1} n_i, \text{where} \ (n_1,...,n_{k-1}) \in T_1 \times ... \times T_{k-1}  \}$
    \State Select the nearest integer design $\mathrm{n}_{\text{init.}} = \arg \min_{n \in T} g(n)$
    \State Calculate the smallest eigenvalue $\lambda$ of $H_g (x^*)$
    \State Calculate radius $r = \sqrt{\frac{2 (g(\mathrm{n}_{\text{init.}}) - g(x^*))}{\lambda}}$
    \For{$i = 1,...,k-1$}
        \State Define $S_i = \{ n_i \in \mathbb{N}: x_i^* \leq n_i \leq \max (x_i^* + r, \eta) \}$
    \EndFor
    \State Define $S = \{ (n_1,...,n_{k-1}, n_k): n_k = \eta - \sum_{i=1}^{k-1} n_i, \text{where} \ (n_1,...,n_{k-1}) \in S_1 \times ... \times S_{k-1}  \}$
    \State Select the integer-optimal design $n^* = \arg \min_{n \in S} g(n)$ by an exhaustive search.
\Ensure $n^*$ 
\end{algorithmic}
\end{algorithm}

\begin{restatable}[Integer-Optimal Design]{thm}{integerOptiamlDesign} \label{thm: integer-optimal Design}
    Algorithm~\ref{alg: integer-optimal design} outputs the integer-optimal design \eqref{eq: integer-optimal design}.
\end{restatable}


\begin{remark}
    Before entering the second for-loop in Algorithm~\ref{alg: integer-optimal design}, the practitioner may decide to check the optimality gap $[g(n_{\text{init.}}) - g(x^*)]/g(x^*)$. If the optimality gap is sufficiently small, it may be acceptable to adopt the suboptimal design $n_{\text{init.}}$. This is demonstrated in Section~\ref{subsec: computation efficiency}.
\end{remark}

\begin{figure}[t]
\hspace{5cm}
\begin{tikzpicture}[line cap=round,line join=round,>=triangle 45,x=1.0cm,y=1.0cm, scale=1.5]
    \clip(-0.3,-0.3) rectangle (4.4,3.3);
    \draw [dotted, rotate around={45.:(1.5,1.7)},line width=1pt, blue] (1.5,1.5) ellipse (1.8cm and 1.2cm);
    \draw [loosely dashed, line width=1pt] (1.4,1.6) circle (1.5cm);
    \draw [dashed, line width=1pt] (1.4,1.6)-- (1.4,3.1);
    \draw [dashed, line width=1pt] (2.9,3.1)-- (-0.1,3.1);
    \draw [dashed, line width=1pt] (2.9,0.1)-- (-0.1,0.1);
    \draw [dashed, line width=1pt] (2.9,0.1)-- (2.9,3.1);
    \draw [dashed, line width=1pt] (-0.1,0.1)-- (-0.1,3.1);
    \draw [line width=2pt, green!70!teal] (2,3)-- (0,3);
    \draw [line width=2pt, green!70!teal] (2,1)-- (0,1);
    \draw [line width=2pt, green!70!teal] (2,1)-- (2,3);
    \draw [line width=2pt, green!70!teal] (0,1)-- (0,3);

    
\begin{scriptsize}
    \draw [fill=black] (0.,0.) circle (1.5pt);
    \draw [fill=black] (0.,1.) circle (1.5pt);
    \draw [fill=black] (0.,2.) circle (1.5pt);
    \draw [fill=black] (0.,3.) circle (1.5pt);
    \draw [fill=black] (1.,3.) circle (1.5pt);
    \draw [fill=black] (2.,3.) circle (1.5pt);
    \draw [fill=black] (3.,3.) circle (1.5pt);
    \draw [fill=black] (3.,2.) circle (1.5pt);
    \draw [fill=black] (3.,1.) circle (1.5pt);
    \draw [fill=black] (3.,0.) circle (1.5pt);
    \draw [fill=black] (2.,0.) circle (1.5pt);
    \draw [fill=black] (1.,0.) circle (1.5pt);
    \draw [fill=black] (1.,1.) circle (1.5pt);
    \draw [fill=black] (1.,2.) circle (1.5pt);
    \draw [fill=black] (2.,2.) circle (1.5pt);
    \draw [fill=black] (2.,1.) circle (1.5pt);
    \draw [fill=black] (1.4,1.6) circle (1.5pt);
    \draw[color=black] (1.6,1.75) node {$x^*$};
    \draw[color=black] (1.8,2.85) node {$n^*$};
    \draw[color=black] (1.1,2.15) node {$n_{\text{inti.}}$};
    \draw[color=black] (1.55,2.25) node {r};
\end{scriptsize}
\end{tikzpicture}
    \caption{Illustration of Algorithm \ref{alg: integer-optimal design}: 
    level set (dotted blue), $\overline{{B_{x^*}(r)}}$ and its enclosing cube (dashed black), feasible region induced from $r$ (thick green), integer points (black)}
    \label{fig: alg}
\end{figure}
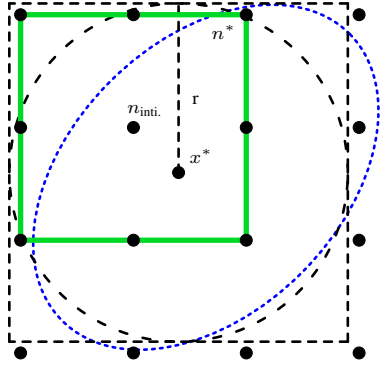

Figure~\ref{fig: alg} illustrates the big picture of Lemma~\ref{lemma: range to search} and Algorithm~\ref{alg: integer-optimal design}. The dotted blue ellipse represent a level set of the variance objective $g$. Our method starts from selecting the nearest integer design $n_{\text{init.}}$ and uses it to establish a ball $\overline{{B_{x^*}(r)}}$ that contains the integer-optimal design with radius $r=\sqrt{2 (g_j (n_{\text{init.}}) - g_j (x^*))/\lambda}$. Then, the solid green cube represents the valid and possible designs for the integer-optimal design, which can potentially be farther from the initial nearest integer design and might not be unique.

\begin{remark}
    Strong convexity of the variance objective plays a key role in identifying the optimal design. While an exhaustive grid search has $\binom{\eta-1}{k-1}$ different combinations, scaling $O(\eta^{k-1})$ when $k$ is fixed, Algorithm~\ref{alg: integer-optimal design} reduces $\eta$ to $2r$, resulting in a much lower complexity of $O\left((2r)^{k-1}\right)$. We show in Section \ref{sec: Simulation} that $r$ is relatively small. 
\end{remark}

\section{Simulation} \label{sec: Simulation}

We numerically illustrate our method through simulation studies. Section~\ref{subsec: cause of naive} compares compares variances between naive and DP-aware stratified sampling. Section~\ref{subsec: interplay between} explores the interplay between the non-private and purely DP designs. Section~\ref{subsec: computation efficiency} showcases the computational efficiency of our algorithm. The input of Algorithm~\ref{alg: integer-optimal design}, $x^*$, is obtained by package \texttt{nloptr} and \texttt{alabama} in \texttt{R}. All computations, including runtime, were conducted on a single-core cluster.

\subsection{Suboptimality of Naive Stratified Subsampling} \label{subsec: cause of naive}

As shown in Table~\ref{tab: variance ratio demo}, stratified sampling under our DP framework requires a tailored design to minimize the variance objective. Both private mean estimation and private A-optimal estimation face variance inflation under specific privacy mechanisms.

In this simulation, there are $4$ groups with population sizes $N=(7000,8000,9000,10000)$ and variance $\sigma^2 = (0.08, 0.08^2, 0.08^3, 0.08^4)$ and a total sample size $\eta=200$. We plot the variance ratio from a naive subsampling scheme to that of the integer-optimal design while varying $\epsilon$ from $0.01$ to $100$.

\begin{figure}[t]
    \centering
    \includegraphics[scale=0.25]{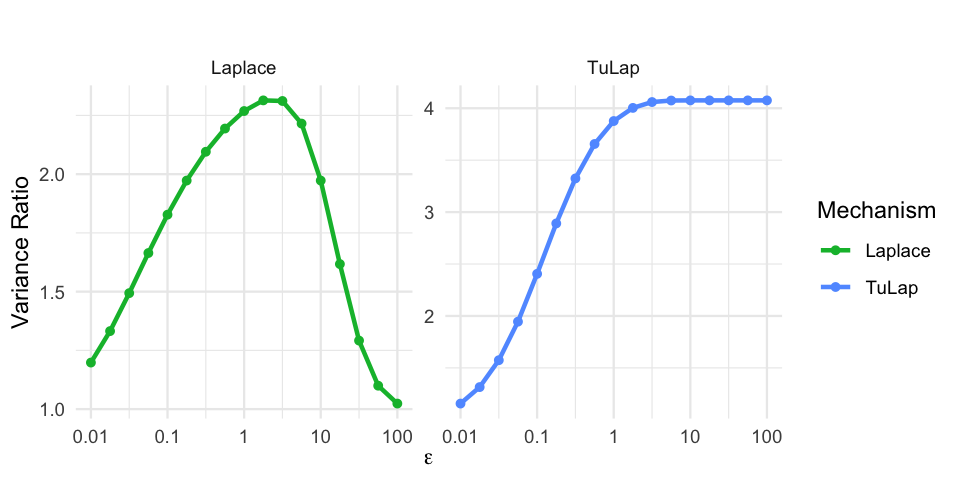}
    \caption{Variance Ratio on population mean}
    \label{fig:Variance Ratio on population mean}
\end{figure}

For the population mean case, Figure~\ref{fig:Variance Ratio on population mean} illustrates that, under the Laplace mechanism, the naive subsampling variance can be up to 2.5 times larger than the optimal design variance within $1<\epsilon<10$. Under TuLap, the variance ratio can reach as high as $4$. Note that DLap gives the same design for population mean.

For the A-optimal case, it reveals a similar trend for the Laplace and TuLap mechanisms as observed in the population mean case (See appendix). 

\subsection{Interplay between No-noise and Purely Laplace Noise Optimal Designs} \label{subsec: interplay between}

In Section~\ref{subsec: closed form solution}, while a closed-form solution for private mean estimation under the Laplace mechanism is unavailable, the optimal design tends to fall between the no-noise and pure-Laplace noise designs.

\begin{figure}[t]
    \centering
    \includegraphics[scale=0.25]{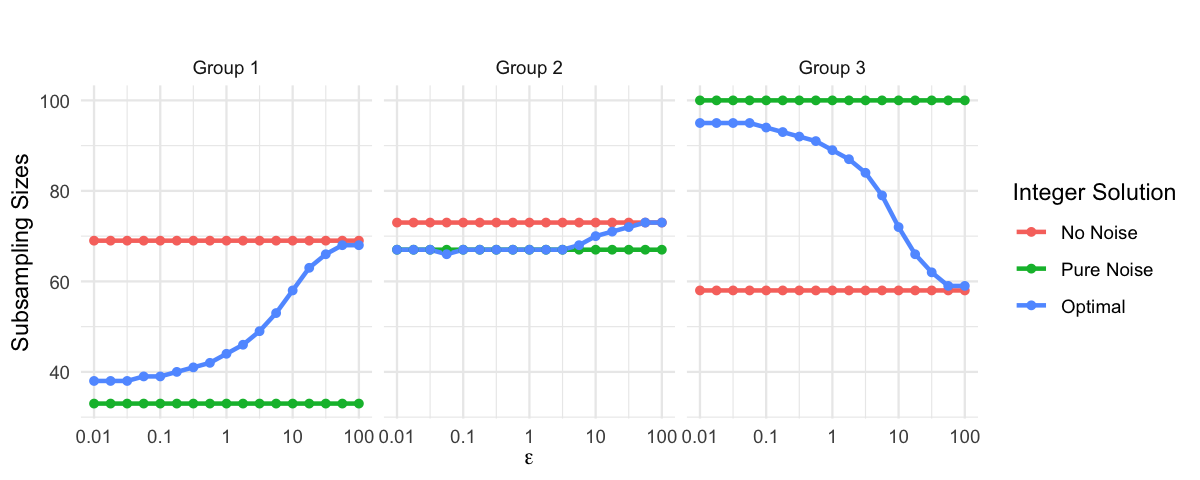}
    \caption{Optimal, no-noise, and pure-noise designs as $\epsilon$ varies.}
    \label{fig:Subsampling Sizes}
\end{figure}

In our simulation, there are $3$ groups with population sizes $N=(1000,2000,3000)$, $\sigma^2 = (0.08, 0.08^{1.5}, 0.08^2)$, $\epsilon=1$ and a total sample size $\eta=200$. We use the setting of the population mean with Laplace noise. Figure~\ref{fig:Subsampling Sizes} demonstrates that the integer-optimal design largely interpolates between the no-noise and pure-noise designs. As privacy protection strengthens, the design shifts closer to the pure-noise configuration; conversely, with weaker privacy protection the optimal design more closely aligns with the Neyman allocation.

\subsection{Computation Efficiency of Algorithm~\ref{alg: integer-optimal design}} \label{subsec: computation efficiency}

Our problem is formulated as a mixed-integer programming task, which lacks an off-the-shelf solution method, particularly in \texttt{R}. While exhaustive search becomes infeasible as either the total sample size $\eta$ or the number of groups $k$ increases, our algorithm proves to be relatively efficient for practical implementation. Although it struggles with scenarios involving large $k$, it remains highly efficient for substantial $\eta$ values—up to $100,000$ and more—when $k$ is kept at a reasonable scale.

\begin{figure}[t]
    \centering
    \includegraphics[scale=0.25]{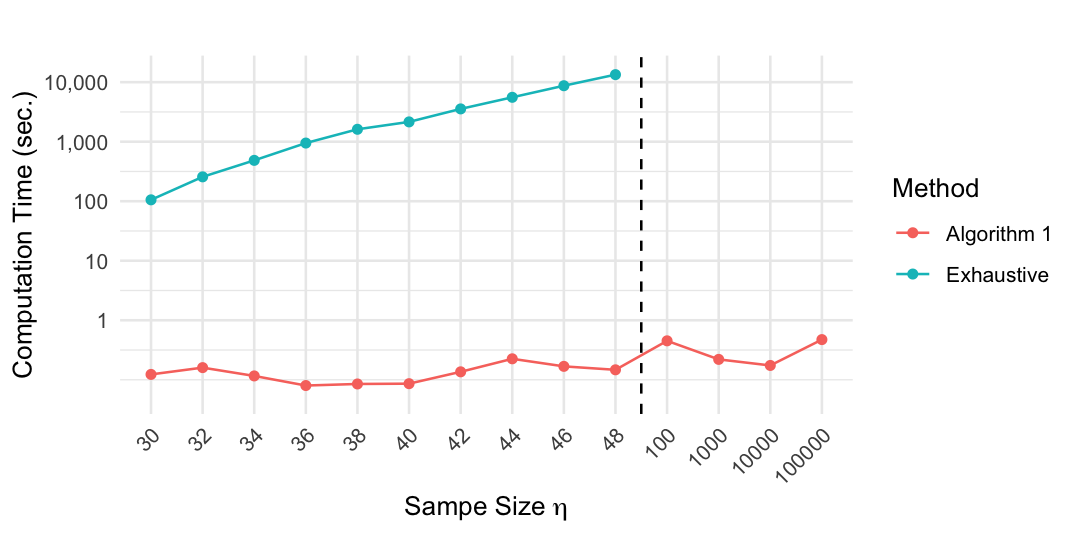}
    \caption{Computation time comparison}
    \label{fig:Computation time comparison}
\end{figure}

In this simulation, there are $10$ groups with $N=(N_1,N_2,\ldots,N_{10}) = (20000, 19000,\ldots,11000)$ and $\sigma^2 = (\sigma^2_1, \sigma^2_2,\ldots,\sigma^2_{10}) = (0.08^{1.1}, 0.08^{1.2},\ldots,0.08^{2})$ and $\epsilon=1$. We measure the computation time for an exhaustive search and for our proposed algorithm as the total sample size $\eta$ increases from $30$ to $48$, using the population mean case with Laplace noise. 
Figure~\ref{fig:Computation time comparison} demonstrates that the exhaustive search exhibits exponential growth in computation time, whereas Algorithm~\ref{alg: integer-optimal design} effectively mitigates this growth. In fact, we see that Algorithm~\ref{alg: integer-optimal design} can effectively find the optimal solution with sample sizes up to $10^5$ in less time than an exhaustive search takes for $\eta=30$.

\begin{figure}[t]
    \centering
    \includegraphics[scale=0.25]{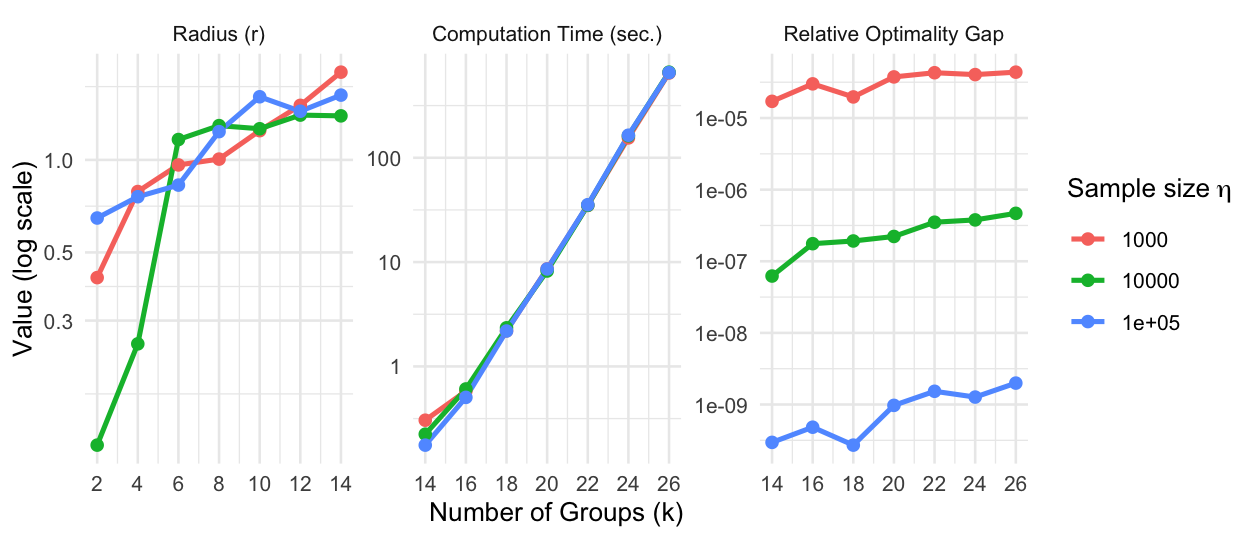}
    \caption{Curse of dimensionality in number of groups}
    \label{fig:Curse of dimensionality in number of groups}
\end{figure}

We consider $k$ groups with $N=(N_1,N_2,\ldots,N_k) = (10000+1000 \cdot k, 10000+900 \cdot k, \ldots, 11000)$ and $\sigma^2 = (\sigma^2_1, \sigma^2_2,\ldots,\sigma^2_k) = (0.08^{1.1}, 0.08^{1.2},\ldots,0.08^{1 + k/10})$ and $\epsilon=1$. We implement our full algorithm from $k=2$ to $12$, locating the optimal design $n^*$, and that of the first part of our algorithm from $k=14$ to $26$, locating the nearest integer design $n_{\text{init.}}$.

If $r>1$, then as $k$ increases the computation time exhibits exponential growth, which becomes large especially when $k \geq 14$. In practice, if $r > 1.5$ and $k$ is large, we recommend identifying the nearest integer design $n_{\text{init.}}$, corresponding to the first half of the algorithm. In this simulation, it maintains an optimality gap of less than $10^{-4}$ from the continuous-optimal design $x^*$. Notably, locating the continuous-optimal design is independent of  Algorithm~\ref{alg: integer-optimal design} and requires less than $1$ second for any $k$ in the simulation.

\section{Discussion} \label{sec: discussion}

We proposed a new framework for integrating differential privacy (DP) into stratified sampling, aiming to achieve two main goals: minimizing the total variance and protecting individual privacy. A significant contribution of our work is the inclusion of privacy considerations in the data collection design. While traditional stratified sampling focuses on minimizing variance assuming non-private data, our approach takes into account the noise introduced by privacy mechanisms as well as the privacy amplification by subsampling, ensuring more reliable results under privacy constraints.

Our framework is fairly flexible, as it works with three common DP mechanisms: Laplace, Discrete Laplace, and Truncated-Uniform-Laplace. Furthermore, the strong convexity of the variance objective ensures that the optimization problem is well-defined, which leads to solid theoretical results. We also addressed the computational challenges of exhaustive search methods by developing an efficient algorithm for finding the optimal integer design. 

However, there are some limitations to our framework. We assume prior knowledge of population variances across groups. In practice, a pilot study is commonly conducted, where a small portion of pilot samples are drawn from each group to estimate group sample variances. The fixed sample size constraint, while reasonable, may be too strict or not account for more complex scenarios (e.g. chance constraints). Additionally, our algorithm may be inefficient when the search radius $r$ is large. Exploring alternative methods for finding the integer-optimal design could improve efficiency; for example \texttt{Gurobi}, \texttt{MOSEK}, and \texttt{Julia}'s \texttt{Pajarito} all have generic mixed-integer programming packages. Lastly, our analysis is limited to $\epsilon$-DP and the Laplace, DLap, and TuLap mechanisms. Future work could extend our results to other mechanisms and privacy frameworks, such as the Staircase mechanism \citet{geng2015optimal}, Gaussian mechanism, or general canonical noise distributions \citet{awan2023canonical} as well as the $\rho$-zCDP \citep{bun2016concentrated}, $\mu$-GDP, and $f$-DP frameworks \citep{dong2022gaussian}.

We also suggest several promising directions for future research. Incorporating the double privacy amplification effect of subsampling and shuffling \citep{Li2023locally} could enhance privacy guarantees, while resulting in a more complex optimization problem. Another option is to apply our framework to a fully central DP context, assuming the presence of a trusted data curator, which would also give a different variance objective. Finally, extending the framework to accommodate more complex sampling designs, such as multi-stage or adaptive sampling, would broaden its applicability to diverse survey scenarios. These potential advancements would further solidify the role of DP-aware stratified sampling in privacy-preserving data collection.


\bibliography{icml2025/main}
\bibliographystyle{plainnat}

\newpage
\appendix
\onecolumn
\section{Proofs and Technical Details}

\begin{lemma} \label{lemma: change of variable y > 0}
    For all $y>0$,
    \begin{align} \label{eq: change of variable y}
    \left(\left(1+\frac{1}{y}\right) \log(1+y) - 2\right)^2 - 1 + \log(1+y) > 0.
    \end{align}
\end{lemma}

\begin{proof}
    First, since we focus on the positive real line ($y > 0$), we can do change of variable by letting $y = e^x - 1$ with domain $x > 0$, then \eqref{eq: change of variable y} becomes
    \begin{align*}
        \left( \frac{x e^x}{e^x - 1} - 2 \right)^2 - 1 + x
        .
    \end{align*}
    We can multiply through by $(e^x -1)^2$ to give us the equivalent of problem of establishing that $h(x)>0$, where
    \begin{align*}
        h(x) \coloneqq (x e^x - 2 e^x + 2)^2 + (x-1) (e^x - 1)^2.
    \end{align*}
    To analyze $h(x)$, we compute its derivatives up to the third order:
    \begin{align*}
    \frac{d h(x)}{d x} =& e^{2x}\left(2x^{2}-4x+3\right)+2e^{x}\left(x-2\right)+1 , \\
    \frac{d^2 h(x)}{d x^2} =& 2e^x\left(e^x\left(2x^2-2x+1\right)+x-1\right) , \\
    \frac{d^3 h(x)}{d x^3} =& 2 x e^x (4 x e^x + 1).
    \end{align*}

    We begin by noting that \( \frac{d^3 h(x)}{d x^3} > 0 \) for all \( x > 0 \), which implies that \( \frac{d^2 h(x)}{d x^2} \) is increasing. Since \( \frac{d^2 h(x)}{d x^2} \big|_{x=0} = 0 \), it follows that \( \frac{d^2 h(x)}{d x^2} > 0 \) for all \( x > 0 \). By the same argument, knowing that \( \frac{d h(x)}{d x} \big|_{x=0} = 0 \), we conclude that \( \frac{d h}{d x} > 0 \) for all \( x > 0 \). Finally, since \( h(0) = 0 \), we establish that \( h(x) > 0 \) for all \( x > 0 \).
\end{proof}

\strongConvexity*

\begin{proof}
    Since we intend to prove that the variance objective is strongly convex on the reals, we substitute $x \in \mathbb{R}_+^k$ for $n$ in the proof for distinction. We will first prove that the variance objective is strongly convex as a function of $x$ (without the constraint $\sum x_i=\eta$) and then verify that they remain strongly convex under the constraint of $\sum_{i=1}^k x_i = \eta$.
    
    Let $g_1(n) = \eqref{stat: general variance from Laplace}$. The Hessian matrix $H_f$ is a diagonal matrix with diagonal entry $(H_{g_1})_{ii}$ as follows:
    \begin{align} \label{eq: Hessian diagonal element from Laplace}
        \frac{\partial^2 g_1}{\partial {x_i}^2} (x_1,\ldots,x_k) = 
        2 \alpha_i^2 \left[\frac{\sigma_i^2}{x_i^3} + \frac{(x_i+c_i)^2 \log^2(1+\frac{c_i}{x_i}) - c_i(4 x_i+3c_i) \log(1+\frac{c_i}{x_i}) + 3c_i^2}{x_i^3 (x_i+c_i)^2 \log^4 (1+\frac{c_i}{x_i})} \right],
    \end{align}
    which is positive for $x_i \in (0, \eta)$ for all $i=1,\ldots,k$, where $c_i = (\exp{(\epsilon / \Delta f)} - 1) N_i$. This is because the first term and the denominator of the second term of \eqref{eq: Hessian diagonal element from Laplace} are clearly positive for all $x_i \in (0, \eta)$ and hence it remains to check the numerator of the second term; by letting $y_i = \frac{c_i}{x_i}$ and completing the squares, we have
    \begin{align*}
        &(x_i+c_i)^2 \log^2(1+\frac{c_i}{x_i}) - c_i(4 x_i+3c_i) \log(1+\frac{c_i}{x_i}) + 3c_i^2 \\
        =& (\frac{c_i}{y_i} + c_i)^2 \log^2(1+y_i) - c_i (4 \frac{c_i}{y_i} + 3 c_i) \log(1 + y_i) + 3 c_i^2 \\
        =&c_i^2 \left[ (1+\frac{1}{y_i})^2 \log^2(1+y_i) - (3 + \frac{4}{y_i})  \log(1+y_i) + 3 \right] \\
        =& c_i^2 \left[ \left((1+\frac{1}{y_i}) \log(1+y_i) - 2\right)^2 - 1 + \log(1+y_i) \right] \\
        >& 0 \quad (\text{Lemma}~\ref{lemma: change of variable y > 0}).
    \end{align*}
    
    Similarly, let $g_2(n) = \eqref{stat: general variance from DLap}$ and $g_3(n) = \eqref{stat: general variance from TuLap}$. The Hessian matrix $H_{g_2}$ is a diagonal matrix with diagonal entry $(H_{g_2})_{ii} = 2 \alpha_i^2 \left(\frac{\sigma_i^2}{x_i^3} \right)$ and the Hessian matrix $H_{g_3}$ is a diagonal matrix with diagonal entry $(H_{g_3})_{ii} = 2 \alpha_i^2 \left(\frac{\sigma_i^2 + \frac{1}{12}}{x_i^3} \right)$, which are both positive  for $x_i \in (0, \eta)$ for all $i=1,\ldots,k$. 
    
    Since $g_j$ is strongly convex with respect to $(x_1,\ldots,x_{k-1}, x_k)$ for $j=1,2,3$, it remains to show that they are strongly convex under the constraint $\eta = \sum_{i=1}^k x_i$. Let $P = \{ (x_1,\ldots,x_k): \eta = \sum x_i \}$ and $a, b \in P$. For $t \in (0,1)$, $g_j (ta + (1-t)b) < t g_j (a) + (1-t) g_j (b)$ because $a, b \in \mathrm{dom}(g_j)$; meanwhile, $ta + (1-t)b \in P$ due to the convexity of the constrained subspace. The conclusion follows.
\end{proof}

\closedFormSolution*

\begin{proof}
Consider 
\begin{align*}
    \min& \frac{1}{(\sum N_i)^2} \left\{ \sum_{i=1}^k \left[ \frac{{N_i}^2}{x_i} \tau_i^2 + 2 \frac{x_i + (\exp{(\epsilon / \Delta f)}-1) N_i}{(\exp{(\epsilon / \Delta f)}-1)^2} \right] \right\} \\
    s.t.& \sum x_i = \eta.
\end{align*}
We can write its Lagrangian as
\begin{align*}
    L(\mathrm{x}, \nu) = \frac{1}{(\sum N_i)^2} \left\{ \sum_{i=1}^k \left[ \frac{{N_i}^2}{x_i} \tau_i^2 + 2 \frac{x_i + (\exp{(\epsilon / \Delta f)}-1) N_i}{(\exp{(\epsilon / \Delta f)}-1)^2} \right] \right\} + \nu (\sum x_i - \eta).
\end{align*}
The KKT condition implies that for all $1\leq i \leq k$,
\begin{equation*}
    \begin{cases}
        \frac{\partial L}{\partial x_i} = \frac{1}{(\sum N_i)^2} \left[ \frac{-{N_i}^2}{{x_i}^2} \tau_i^2 + \frac{2}{(\exp{(\epsilon / \Delta f)}-1)^2} \right] + \nu = 0 \\
        \frac{\partial L}{\partial \nu} = \sum x_i - \eta = 0.
    \end{cases}    
\end{equation*}

Then, $\nu (\sum N_i)^2 + \frac{2}{(\exp{(\epsilon / \Delta f)}-1)^2} = \frac{{N_i}^2}{{x_i}^2} \tau_i^2$ implies $x_i \propto \tau_i N_i$, as the LHS of the equation is constant for all $i$. Therefore,
\begin{align*}
    x_i^* = \frac{\tau_i N_i}{\sum \tau_i N_i} \eta, 
\end{align*}
where $\tau_i^2 = \sigma_i^2$ for Discrete Laplace and $\tau_i^2 = \sigma_i^2 + \frac{1}{12}$ for TuLap. 
\end{proof}

\closedFormSolutionPureDP*

\begin{proof}
We can express the problem as an equality-constrained minimization problem
\begin{align*}
    \min& \frac{\sum N_i \varphi(\frac{x_i}{N_i})}{(\sum N_i)^2}\\
    s.t.& \sum x_i = \eta,
\end{align*}
where, with $c = \exp{(\epsilon / \Delta f)}-1$, $\varphi (y) = \frac{2}{y} \log^{-2} \left( 1+ \frac{c}{y} \right)$ is strongly convex w.r.t. $y$. Its Lagrangian is
\begin{align*}
    L(x, \nu) = \frac{\sum N_i \varphi (\frac{x_i}{N_i})}{(\sum N_i)^2} + \nu (\sum x_i - \eta).
\end{align*}
The KKT condition implies that for all $ 1\leq i \leq k$
\begin{equation*}
    \begin{cases}
        \frac{\partial L}{\partial x_i} = \frac{ \varphi'(\frac{x_i}{N_i}) }{(\sum N_i)^2} + \nu = 0 \\
        \frac{\partial L}{\partial \nu} = \sum x_i - \eta = 0.
    \end{cases}
\end{equation*}

Thus, $-\nu \sum N_i = \varphi'(\frac{x_1}{N_1}) = \cdots = \varphi'(\frac{x_k}{N_k})$. Note that $\varphi$ is strongly convex if and only if $\varphi'$ is strictly increasing, which implies $\frac{x_1}{N_1} = \frac{x_2}{N_2} = \cdots = \frac{x_k}{N_k}$ is a sufficient and necessary condition for the system of equations to hold. Now, since $\sum x_i = \eta$ and $x_i \propto N_i$ (i.e. $x_i = b N_i$ for some $b \in \mathbb{R}$), $\sum x_i = b \sum N_i = \eta$ and hence $b = \frac{\eta}{\sum N_i}$, which gives
\begin{equation*}
    \begin{cases}
        x_i^* = \frac{\eta}{\sum N_i} N_i \ \text{ for all }i, \\
        \nu^* = \frac{-1}{(\sum N_i)^2} \varphi'(\frac{\eta}{\sum N_i}).
    \end{cases}
\end{equation*}
\end{proof}

\section{Supplementary Graph for Simulations}

\begin{figure}[H]
    \centering
    \includegraphics[scale=0.23]{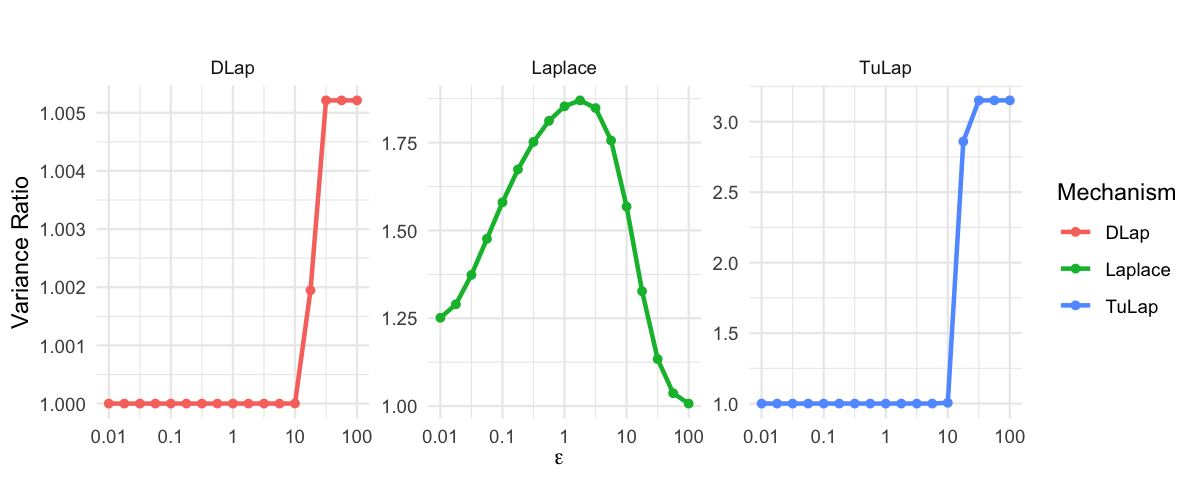}
    \caption{Variance Ratio for A-Optimal Design}
    \label{fig:Variance Ratio on A-Optimal}
\end{figure}

For A-optimal cases, illustrated in Figure~\ref{fig:Variance Ratio on A-Optimal}, the naive subsampling variance under the Laplace mechanism can be up to 1.8 times larger than the optimal design variance within the range $1 < \epsilon < 10$. For the TuLap mechanism, a noticeable jump in the variance ratio occurs around $\epsilon = 10$, driven by the variance of the Unif$(0,1)$ component. Similarly, the DLap mechanism follows a trend comparable to TuLap but demonstrates the smallest variance ratio gap between the naive subsampling scheme and the optimal integer design. This behavior aligns with the population mean scenario, where the ratio remains constant at $1$. 

\end{document}